\documentclass[journal]{IEEEtran}
\usepackage[utf8]{inputenc} % allow utf-8 input
\usepackage[T1]{fontenc}    % use 8-bit T1 fonts
\usepackage{graphicx}%
\usepackage{multirow}%
\usepackage{amsmath,amssymb,amsfonts}%
\usepackage{amsthm}%
\usepackage{mathrsfs}%
\usepackage{xcolor}%
\usepackage{textcomp}%
\usepackage{manyfoot}%
\usepackage{booktabs}%
\usepackage{algorithm}%
\usepackage{algorithmicx}%
\usepackage{algpseudocode}%
\usepackage{listings}%
\usepackage{subcaption}
\usepackage{tabularx}
\usepackage{multirow}  % for \multirow
\usepackage{multicol}
\usepackage{booktabs} 
\usepackage{hyperref}       % hyperlinks
\usepackage{url}  
\usepackage{nicefrac}       % compact symbols for 1/2, etc.
\usepackage{microtype}      % microtypography
\usepackage{float}
\usepackage{array}
\begin{document}

\title{Self-contained Data Cleaning for Unsupervised Anomaly Detection}


\author{Jordan F. Masakuna$^{1,2}$, Ahmed Yacine$^{1}$, Djeff K. Nkashama$^{1}$, Arian Soltani$^{1}$, Marc Frappier$^{1}$, Pierre M. Tardif$^{1}$ and Froduald Kabanza$^{1}$\\
\IEEEauthorblockA{$^1$University of Sherbrooke, Sherbrooke, Canada\\}
\IEEEauthorblockA{$^2$University of Kinshasa, Kinshasa, DR. Congo\\}
        % <-this % stops a space
%\thanks{This paper was produced by the IEEE Publication Technology Group. They are in Piscataway, NJ.}% <-this % stops a space
}

\maketitle
\begin{abstract}
Training datasets for unsupervised anomaly detection are often assumed to be anomaly-free, but in practice, they frequently contain malicious or anomalous contamination. This contamination severely degrades the dependability and security of downstream detection models. While some contamination-robust methods exist, they rarely generalize across different model architectures and typically rely on a priori knowledge of the contamination ratio\textemdash information that is rarely available in real-world security operations. In this paper, we propose a novel, self-contained preprocessing framework for anomaly detection that identifies and removes training set contamination without relying on known contamination ratios. Our approach performs a dynamic analysis of the cosine similarity between each data point and its reconstruction during the early training stages of an auto-encoder, mapping points into a dynamic space where anomalous regions naturally exhibit lower density. By executing ``hard cleaning''\textemdash removing all identified anomalous candidates from the training pool\textemdash we systematically improve data quality. Because this cleaning occurs at the data level, the security and reliability improvements generalize seamlessly across various downstream anomaly detection architectures. Evaluating our framework across 8 distinct models and 10 security-relevant datasets, our approach achieves up to 100\% contamination reduction, yielding average improvements of 11.2\% in F1 score and 7.3\% in classification accuracy.
\end{abstract}

\begin{IEEEkeywords}
Unsupervised anomaly detection, data contamination, robustness
\end{IEEEkeywords}


\section{Introduction}
\label{sec:introduction}
In the ever-expanding landscape of deep learning (DL), engineering robust and dependable models has become a fundamental imperative for critical infrastructure \cite{masakuna2024streamlined}. DL architectures have demonstrated exceptional capabilities in high-stakes environments, ranging from autonomous vehicles and medical diagnostics to industrial control systems and cyber-threat monitoring. However, as these technologies integrate into dependable system pipelines, their operational reliability remains highly sensitive to the quality of their underlying training data. This vulnerability is commonly characterized by the classic data-quality paradigm where degraded input fidelity directly translates to compromised system outputs \cite{mazumder2024dataperf}. In secure systems, data quality issues encompass not only benign data degradation (e.g., sensor anomalies, noise, and out-of-distribution (OoD) instances) but also malicious data manipulation.

To deploy effective unsupervised anomaly detection systems\textemdash such as network intrusion detection systems (NIDS) or fraud detection engines\textemdash practitioners typically rely on the assumption that an anomaly-free training dataset is accessible. This clean baseline allows the model to map normal system behavior accurately. In real-world security operations, however, this assumption is frequently violated due to data contamination. Contamination may manifest as unintentional faults (such as calibration errors, telemetry glitches, or labeling noise \cite{da2021robust}) or as intentional threats (such as stealthy training data poisoning designed by an adversary to corrupt the model’s decision boundaries). In operational settings, malicious anomalies and normal system behaviors (inliers) often exhibit high structural similarity, making their manual decoupling exceptionally difficult. Ultimately, to guarantee the reliability, availability, and integrity of downstream security operations, anomaly detection architectures must exhibit inherent resilience to contaminated training environments.

Existing defense mechanisms against data contamination are predominantly model-centric, modifying the learning objective or loss function of specific neural network architectures to downweight anomalous samples. However, these techniques suffer from two critical limitations that hinder their integration into dependable systems. First, they lack architectural generalizability, forcing security operators to re-engineer defenses when migrating across different model types. Second, they rely heavily on a priori knowledge of the contamination ratio\textemdash a metric that is practically impossible to estimate accurately in evolving, untrusted environments \cite{perini2023estimating}. In contrast, establishing dependability at the data layer through data-centric security preprocessing offers a highly generalizable and robust alternative \cite{mazumder2024dataperf, pleiss2020identifying}. By identifying and removing anomalous data points prior to training, any downstream detection model can be safely trained on a verified, high-integrity dataset.

\textbf{Contributions}. In this paper, we propose a novel, self-contained data-centric preprocessing framework designed to secure unsupervised anomaly detection pipelines against training data contamination without requiring prior knowledge of the contamination ratio. Our methodology performs a dynamic analysis of the reconstruction error\textemdash specifically utilizing the cosine similarity between a data point and its auto-encoder reconstruction\textemdash during the initial, highly informative epochs of training. By mapping these trajectories, we construct a ``dynamic representation space'' where contaminated samples naturally isolate in low-density regions. Through a process of ``hard cleaning,'' we prune these anomalous candidates from the training pool.

The primary contributions of this work are summarized as follows:
\begin{itemize}
    \item 
\textbf{Model-agnostic dependability}. We present a self-contained data-cleaning framework that executes at the preprocessing stage, ensuring that the resulting dataset improves the robustness of any downstream unsupervised anomaly classification architecture.

\item \textbf{Zero-knowledge thresholding}. Our approach eliminates the dependency on known contamination ratios, making it uniquely suited for real-world security environments where the volume of malicious injection is unknown.

\item \textbf{Extensive empirical validation}. Evaluating our framework across 8 distinct models and 10 multi-domain security and telemetry datasets, we achieve up to 100\% contamination reduction. This translates to substantial performance gains, yielding an average improvement of 11.2\% in F1 score and 7.3\% in classification accuracy.

\item \textbf{Resource efficiency}. By cleaning the dataset beforehand, we eliminate the need for complex, computationally intensive robust-training routines. This allows high-performance anomaly detection to run on low-resource local hardware, lowering deployment costs for edge networks and resource-constrained environments.
\end{itemize}
To define our system boundary, we focus strictly on training data contamination (i.e., identifying and removing anomalous faults present within the offline training set). We do not address test-time adversarial evasion attacks designed to bypass an already-trained model, which represents a distinct security threat model. Additionally, while interpretability is crucial for modern trust in AI, this work focuses exclusively on classification reliability and data integrity.

While we acknowledge that rare edge cases or novel operational patterns may occasionally carry useful semantic features, our system design operates under the standard dependability assumption that training anomalies represent harmful system noise. Consequently, filtering out these anomalous deviations is treated as a necessary step to maximize the precision and recall of the downstream classifier.

\section{Background and related work}
\label{sec:related_work}
\subsection{Background}
In unsupervised anomaly detection, models are designed to capture the underlying boundary of normal system behavior (inliers) to identify deviations that may signify faults, intrusions, or failures \cite{ruff2018deep}. However, the presence of training-phase contamination fundamentally violates this paradigm. When anomalous or malicious data points pollute the training set, they distort the model's representation of the benign distribution, skewing the learned decision boundaries in unpredictable directions. Consequently, the model's ability to distinguish between authorized system states and actual threats is severely compromised. In operational security pipelines, this representational drift manifests as high false-negative rates (unidentified threats) or elevated false-alarm rates, directly undermining the availability and dependability of the detection system.

Early methodologies for behavior profiling and anomaly detection relied heavily on statistical heuristics and distance metrics, such as the Mahalanobis distance and robust covariance estimators \cite{rousseeuw2018anomaly}. While mathematically rigorous, these threshold-based techniques assume specific parametric data distributions, making them highly susceptible to performance degradation when deployed against complex, high-dimensional, and non-linear security telemetry.

To address these distribution constraints, researchers widely adopted probabilistic frameworks, including Gaussian Mixture Models and Bayesian Neural Networks. By modeling uncertainty explicitly, these approaches assign a probability score to incoming system states, facilitating the identification of outliers. However, these frameworks exhibit significant computational overhead and scale poorly to high-velocity, real-world network traffic or enterprise system logs, limiting their utility in real-time intrusion-tolerant systems.

To overcome the scalability and representational bottlenecks of statistical models, modern architectures leverage deep learning DL representations, utilizing frameworks such as auto-encoders and Generative Adversarial Networks (GANs) to perform complex, non-linear boundary learning \cite{xia2022gan}. While DL models excel at learning robust features from raw telemetry, their massive parametric capacity makes them highly sensitive to training-set contamination, as they easily memorize anomalous patterns during training.

To mitigate generalization limits, security operators have explored data augmentation techniques adapted for anomaly detection \cite{lu2021gan}. By synthesizing diverse representations of normal operational data, these methods attempt to construct a more resilient normal boundary. However, while data augmentation can broaden the coverage of benign profiles, it fails to address\textemdash and can inadvertently propagate\textemdash inherent contamination present within the seed training dataset.

Recognizing this limitation, emerging research in data-centric security emphasizes systemic data cleaning prior to model training \cite{mazumder2024dataperf}. This paradigm recognizes that the integrity of the downstream classifier is bounded by the purity of the training data. Rather than relying on complex, model-specific robust loss formulations, data-centric cleaning establishes a verified, high-integrity baseline. This approach systematically prevents training faults from degrading the model, ensuring dependable downstream performance across arbitrary neural network architectures.
\subsection{Related work}
\subsubsection{Dependability}
Unsupervised anomaly detection serves as a vital behavioral profiling mechanism for preserving the dependability, availability, and integrity of critical systems, such as NIDS and industrial telemetry networks. However, the standard assumption that training datasets are entirely benign is routinely violated in real-world deployments due to either accidental system faults (e.g., sensor telemetry glitches) or malicious adversarial actions (e.g., training data poisoning) \cite{da2021robust}. When training sets are contaminated, the integrity of downstream machine learning models is severely compromised, distorting learned decision boundaries and leading to catastrophic system-level failures during inference \cite{ruff2018deep}.

To mitigate the risks of training-phase contamination, existing research has heavily focused on model-centric techniques designed to limit the influence of anomalous samples during model optimization. For instance, robust deep auto-encoders and optimization frameworks utilize custom loss formulations or adaptive sample weighting to downweight high-reconstruction-error instances during the training loop. 

While theoretically appealing, these model-centric approaches present two critical bottlenecks that limit their integration into dependable systems:
\begin{enumerate}
    \item \textbf{Architectural dependency.} Because these defense mechanisms are integrated directly into the loss functions of specific neural network architectures, they do not generalize across different models. A system operator migrating from a dense auto-encoder to a convolutional or graph-based model is forced to entirely redesign and retrain the robust defense mechanism.
    \item \textbf{Contamination ratio dependency.} The vast majority of robust models rely on prior knowledge of the exact contamination ratio within the training dataset to establish pruning thresholds or anomaly scoring baselines. In real-world security operations, this ratio is a zero-knowledge variable. Relying on system operators to accurately estimate the proportion of stealthy, actively injected anomalies introduces a severe point of failure.
\end{enumerate}

To establish baseline profiles of normal operations, historical systems relied on classical statistical estimators, such as the Minimum Covariance Determinant and robust Mahalanobis distance thresholds \cite{rousseeuw2018anomaly}. While mathematically rigorous, these parametric methods scale poorly when faced with highly non-linear, high-dimensional security telemetry (e.g., raw network flow data) and suffer from high computational complexity.

To bypass these limitations, modern deep learning architectures\textemdash specifically DAEs and GANs\textemdash have been widely adopted due to their high representational capacity \cite{xia2022gan}. DAEs learn a low-dimensional bottleneck representation of normal traffic, flagging inputs with high reconstruction errors as anomalous. Yet, because of their massive capacity, standard DAEs suffer from excessive generalization; if trained on contaminated datasets, they readily reconstruct and memorize the anomalous samples. This representational bleed causes the model to classify active threats as benign during production, resulting in catastrophic security evasion.

The fundamental limitations of model-centric robustness have catalyzed a paradigm shift toward data-centric security \cite{mazumder2024dataperf}. Rather than adapting complex, resource-heavy architectures to withstand noisy environments, data-centric systems focus systematically on preprocessing steps that isolate and remove faults at the data level before downstream modeling begins \cite{pleiss2020identifying}. Existing preprocessing techniques include data cleaning, sample pruning, and noise estimation thresholds. 

While highly effective for general machine learning tasks, existing data-cleaning algorithms still fail when applied to unsupervised anomaly detection:
\begin{itemize}
    \item \textbf{Data augmentation failures.} Methods that use generative networks to augment normal data profiles help models generalize better, but they do not actively purge malicious data points from the initial training set \cite{lu2021gan}.
    \item \textbf{Computational latency.} Interleaved clustering and representation-learning approaches (e.g., Iterative Data Distillation) provide a route to data selection but suffer from extreme computational complexity. This makes them impractical for low-resource edge servers or startups with strict operational constraints.
\end{itemize}

Our work addresses these interconnected limitations by proposing a self-contained, data-centric preprocessing approach focusing on ``hard cleaning.'' Instead of relying on complex, architecture-dependent loss adjustments or assuming prior knowledge of the contamination ratio, we exploit the natural training dynamics of deep networks. By analyzing the trajectory of the cosine similarity between data points and their reconstructions in the earliest epochs of auto-encoder training, we map the training set into a dynamic space. 

Because anomalous regions naturally map to low-density zones in this dynamic space during early training, our framework systematically isolates and prunes them. By removing training-phase faults \textit{prior} to downstream model training, we deliver a guaranteed high-integrity, clean dataset. This model-agnostic protection ensures that any standard, resource-efficient unsupervised classifier subsequently trained on the cleaned data inherits robust, dependable performance without incurring the heavy computational overhead of robust training.
\subsubsection{Deep learning}
Two categories of contamination-resilient methods are related to our preprocessing anomaly detection approach: one pertains to unsupervised  anomaly detection, while the other concerns supervised techniques, primarily applied to handle mislabelled images\textemdash noisy label issue in multi-class image classification is related to data contamination in the context of unsupervised anomaly detection. 

\textbf{Unsupervised anomaly detection approaches for contaminated data}.  An interesting contamination-robust technique (called NeutrALAD) was introduced in \cite{qiu2022latent}. It involves learning labels (inlier vs. anomaly) of data points during training, to limit the impact of potential anomalies in the objective function. Its ability to generalize to unseen data and learn more representative features makes it versatile across various domains and data types. However, NeutrALAD is complex and requires careful hyperparameter tuning, which can be resource-intensive. Additionally, it relies heavily on the initial quality of the training data and may lack interpretability, making it difficult to understand its decision-making process.
    Another robust unsupervised technique (called DUAD) was introduced in \cite{li2021deep}. DUAD involves iteratively searching for a less contaminated subset of a contaminated data set, which, however, makes it inefficient in terms of complexity. Its architecture allows for the robust learning of feature representations, enhancing its ability to detect subtle and complex anomalies without requiring labeled training data.
    Nonetheless, NeutrALAD and DUAD require information on the  contamination ratio.
   Though still in its early stage, estimating the contamination ratio if unavailable is a research question on its own (e.g., \cite{perini2023estimating}), which might form prior work to train models such as DUAD. %But the method in \cite{perini2023estimating} is  inefficient in terms of complexity as it is a fusion method and provides inaccurate contamination ratios in most cases we consider. 
   \cite{kim2023iterative} proposed a contamination-robust framework called IAD, implemented on three types of anomaly detection methods (probabilistic, one-class classification and reconstruction-based methods).  IAD involves quantifying weights associated to data points during training  so that candidates with low weights would have less impact in the objective function, all without relying on the contamination ratio. Weights are computed solely based on reconstruction errors during training. 
   \\
    \textbf{Contrast}. 
    Our work introduces a self-contained preprocessing anomaly detection approach that prioritizes data cleaning in scenarios where information on contamination ratios is unavailable\textemdash such information may not always be readily accessible due to factors like the absence of ground truth for anomalies \cite{kim2023iterative}. Most notably, our preprocessing technique yields an ``cleaned`` data set (i.e., data sets with less contamination) suitable for training any unsupervised anomaly detection model that does not rely on contamination ratio information, including IAD framework.
    
   \textbf{Multi-class supervised approaches for noisy labels}. 
   Multi-class classification strategies \cite{masakuna2020performance} for handling noisy labels have attracted significant attention, particularly in the domain of image classification. Noteworthy among these is the AUM method \cite{pleiss2020identifying}, which delves into gradient dynamics to discern signals indicative of mislabeled images. Similarly, MarginGAN \cite{dong2019margingan} explores margin maximization concepts to achieve similar objectives.
Confident learning methodologies have emerged as effective means to enhance label quality in noisy data sets \cite{northcutt2021confident}. By identifying and prioritizing confident samples\textemdash those characterized by low loss values and large margin rankings\textemdash these techniques mitigate the impact of noisy data. They involve a range of strategies, including the pruning of noisy data, the establishment of probabilistic thresholds for noise estimation, and the ranking of examples for confident training.
Moreover, addressing noisy labeling remains a prominent research endeavor, with various strategies proposed to alleviate its effects. Techniques such as consistency-based augmentation \cite{reed2014training} and label propagation algorithms \cite{wang2006label} offer promising avenues for managing noisy labels and improving classification performance.
In parallel, there has been notable progress in learning from noisy data sets with minimal supervision. Leveraging a combination of large data sets with noisy annotations and smaller sets of cleanly annotated data, these methods aim to derive robust representations capable of handling noise effectively.
Furthermore, advancements in loss function design have been pivotal in enhancing discrimination power within deep neural networks. By introducing novel loss functions tailored to maximize inter-class variance while minimizing intra-class variance, these approaches offer more effective solutions to noisy label issues.
\\
 \textbf{Contrast}.
 We intend to adapt the concept of the dynamics of gradient (e.g., AUM) on cosine similarity as a preprocessing step for unsupervised anomaly detection in order to handle data contamination. We expect anomalies to be sparse in the dynamic space. We will examine the dynamics using standard deviation of changes and  apply a classical anomaly detection method on the dynamic space to identify potential anomalies. We will employ an auto-encoder (which we briefly introduce next) to generate the dynamics of cosine similarity (dynamic space) between each point and its reconstructions at training.

%Unsupervised anomaly detection is primarily geared towards uncovering the inherent structure of inliers. However, when anomalies infiltrate the training data set, distinguishing between inliers and outliers becomes challenging. In such scenarios, the model (an auto-encoder is used) erroneously learn all instances as inliers. A promising approach to identify anomalies involves analyzing the dynamics of the loss function at each instance. We anticipate that anomalous instances will exhibit irregular behavior, facilitating their recognition. 

\textbf{Auto-encoder (AE)}.
A contaminated training data set, denoted $X=X^+\cup X^-$, contains some anomalies (where $X^+$ and $X^-$ denote subsets of inliers and anomalies respectively).
Let $h_\Theta^{(t)}$ denote an AE, $\hat{x}_i^{(t)}$ the reconstructed version of $x_i$ and $e_i^{(t)}$ the reconstruction error on $x_i$ at training iteration $t$.  The reconstruction error $e_i^{(t)}$  is given by
$e_i^{(t)} = \|x_i-\hat{x}_i^{(t)}\|\,,\text{ where }\,\hat{x}_i^{(t)} =h_\Theta^{(t)}(x_i)\,.$
The loss function, using an AE, $h_\Theta$, trained on $X$, is 
\begin{equation}
\label{eq:loss}
    L(\Theta; X)=\sum_{x_i\in X} \|x_i-\hat{x}_i^{(t)}\|^2\,.
\end{equation}
We opt for the $\ell_2$-norm in \eqref{eq:loss}  due to its anomaly sensitivity, enhancing  characterization of anomalies.

 \begin{figure}[h]
    \centering
        \begin{subfigure}[b]{0.15\textwidth}
        \includegraphics[width=\textwidth]{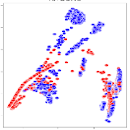}
        \caption{}
        \label{fig:initial}
    \end{subfigure}
      \begin{subfigure}[b]{0.15\textwidth}
        \includegraphics[width=\textwidth]{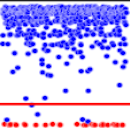}
        \caption{}
        \label{fig:distr2}
    \end{subfigure}
    \caption{Blue and red inliers and anomalies respectively. (\ref{fig:initial}): original vs (\ref{fig:distr2})  dynamic space.}
    \label{fig:illustration}
\end{figure}
AEs  leverage their ability to learn a compact representation of  normality. Essentially, an AE learns to compress and  reconstruct data, becoming efficient at handling similar data. When faced with anomalies, AE tends to reconstruct poorly, as the new data contains patterns not captured during training. This principle underlies the AE's utility in anomaly detection.
However, when trained on contaminated data, AEs may not inherently discriminate between inliers and anomalies based solely on individual loss values. 
Interestingly, a notable pattern has emerged: when we train an AE on contaminated data, most anomalies are found in low-dense regions of the dynamic space. This is depicted in Figure \ref{fig:illustration}. %: \ref{fig:initial} illustrates the training data, where it is difficult to isolate anomalies. In contrast, Figure \ref{fig:distr2} demonstrates a better separation  when plotted in the dynamic space.

It is worth noting that any AE architecture can be used for this step. In our experiments, we intentionally adopt a simple AE to keep the emphasis on the behavior of the dynamic sequences rather than on architectural complexity. Nevertheless, we also evaluate several alternative AE architectures to demonstrate that the results are robust and largely independent of the specific AE design.
\subsection{Formal problem statement for the data cleaning process}

Given a contaminated dataset $X = \{x_1, x_2, \dots, x_n\}$ where some data points are anomalies, a data cleaner function $h_\Theta: X \to X$ that removes anomalies and a cleaned data set $\hat{X} = h_\Theta(X)$, used for unsupervised anomaly detection.
Let  $c: \mathcal{X} \to \{0,1\}$ be a contamination function where
$$
  c(x) =
\begin{cases} 
1, & \text{if } x \text{ is an anomaly (contamination)} \\ 
0, & \text{otherwise}
\end{cases}  
$$
The contaminated dataset
$X = X^{-} \cup X^{+}$ with
\begin{itemize}
    \item $X^{-} = \{ x \in X \mid c(x) = 0 \}$,
    \item $X^{+} = \{ x \in X \mid c(x) = 1 \}$\,.
\end{itemize}
An ideal data cleaner $h_\Theta$ should imply $\hat{X} = h_\Theta(X) = X \setminus X^{+}$ where $X^{+} = \{ x \in X \mid d(x) > \tau \}$ with $d(x)$ being an anomaly scoring function (e.g., cosine similarity distance, reconstruction error, isolation score), and 
$\tau$ being a threshold to classify an instance as an anomaly.
Once $\hat{X}$ is obtained, it is used as input to an unsupervised anomaly detection model $u: \hat{X} \to \{0,1\}$ or $y = u(\hat{X})$ where

$$
u(x) =
\begin{cases} 
1, & \text{if } x \text{ is an anomaly (detected)}\,, \\ 
0, & \text{otherwise}\,.
\end{cases}
$$

This ensures that the data set first undergoes cleaning, removing as many anomalies as possible, and is then passed to an unsupervised anomaly detection method that learns the underlying distribution of the remaining samples. Our augment is that $u: \hat{X} \to \{0, 1\}$ will outperform $u: X \to \{0, 1\}$, i.e., performing unsupervised anomaly detection on the cleaned dataset $\hat{X}$ (after applying the data cleaning function $h_\Theta$) will lead to better classification results than directly applying the detection method on the contaminated dataset $X$. In other words, let $\pi$ denote a performance function,
   $$\pi(u(\hat{X})) \geq \pi(u(X))\,.$$

\section{Preprocessing anomaly detection}
\label{sec:add-on}
\begin{figure}[h]
 \centering
\includegraphics[width=6cm]{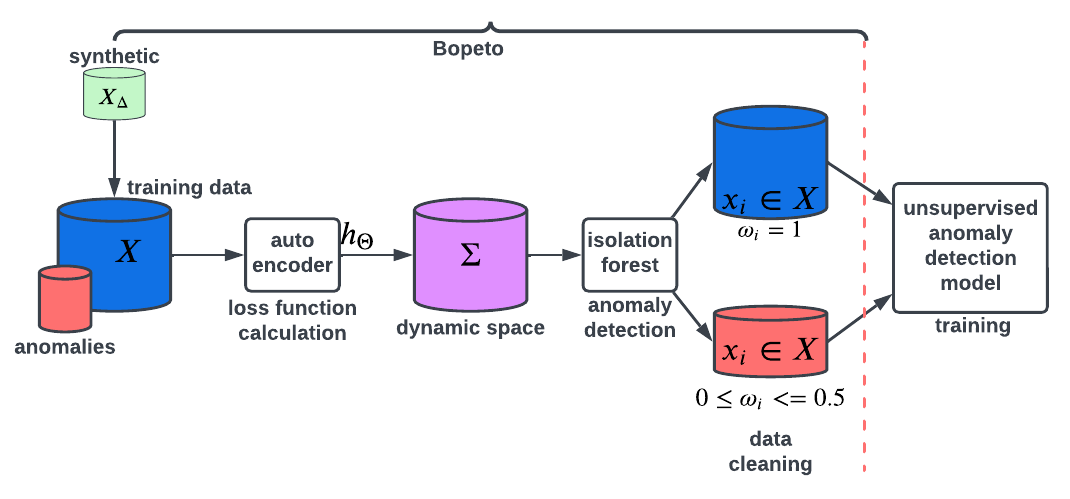}
\caption{Bopeto's architecture.}
\label{fig:archi}
\end{figure}
Our approach (named Bopeto\footnote{Bopeto means hygiene in Lingala, a local language spoken in the Democratic Republic of Cong.}) is data-centric: it identifies as much anomalies as possible in the training  set before being used to build various anomaly detection models that do not explicitly account for contamination ratio.

Bopeto (Figure \ref{fig:archi}) comprises two primary steps: (1) determination of a dynamic space where identifying anomalies becomes more feasible, and (2) anomaly identification and treatment through assigning weights to data points (less weighted are more anomalous). Though we train our model using $\ell_2$-norm (as in \eqref{eq:loss}), we use cosine similarity to evaluate  values that will constitute our dynamics since it is  affected by the direction of vectors. This means that even if the vectors have different scales, their cosine similarity can still be meaningful. Dynamics represent the trajectories of the cosine similarity between a data point and its reconstructions at different training iterations. Subsequently, we utilize a traditional unsupervised anomaly detection method  to pinpoint potential anomalies from the dynamic space. Isolation forest (IF) is one option among many estimators (e.g. density-based clustering and one-class SVM) to cluster dynamics scores into two regions in order to calculate decision threshold that separates normal instances from malicious ones. % within a contaminated data set. %The advantage of IF is that no assumption of data distribution is required and it has a relatively fast inference time. 
%A preliminary step of Bopeto involves introducing synthetic examples into the data set to enhance characterization of anomalies. To generate synthetic data, we employ the fast gradient method (FGM) \cite{goodfellow2014explaining} (which we introduce describe next), which produces inadequately reconstructed and noisy instances (or adversarial examples). Inputs for FGM are partly from within the training data set and partly from its boundaries. 
%\subsection{Fast Gradient Method (FGM)}
%FGM is a technique for generating adversarial examples. It creates perturbations designed to mislead models in a controlled way. 
%FGM perturbs an input data point by using the gradients of the loss function with respect to the input itself. Our goal here is to slightly modify the reconstructed input data to make anomalies more distinctive using an auto-encoder (AE). Here are the steps: (1) choose a reconstructed version $\hat{x}$ of an input $x\in X$; (2) compute the gradient of the loss function with respect to the input $\nabla_x J(\Theta, \hat{x}, \hat{x})$, where $\Theta$ represents AE's parameters.
% (3) generate a perturbation $\gamma$ by applying the sign of the gradient to the input, $\gamma = \epsilon \times \text{sign}(\nabla_x J(\theta, \hat{x}, \hat{x}))\,,$
  % where $\epsilon>0$. %; and (4) construct an adversarial example $x' = \hat{x} + \gamma\,.$

It should be noted that in Bopeto, the autoencoder is deliberately not fully trained. Our analysis is confined to the first ten training iterations, during which the model remains unstable and reconstruction outputs fluctuate substantially. Bopeto leverages this early-stage instability by capturing dynamic reconstruction trajectories that characterize each data point’s response to rapid changes in the learned mapping. Consequently, performance does not depend on the autoencoder’s final accuracy, but on the evolution of its learning dynamics in early iterations, where normal and anomalous samples become naturally separable.
\subsection{Determination of the dynamic space}
Let $s^{(t)}_i=\text{cosine}(x_i, \hat{x}_i^{(t)})$ and $\delta_i^{(t)} = s_i^{(t)}-s_i^{(t-1)}$ denote  the cosine similarity between an input and its reconstructed version at iteration $t$, and the discrepancy between two consecutive cosine similarities on a data point $x_i$ after iteration $t$ respectively, and $T$ the maximum number of training iterations. The dynamics of cosine similarities on $x_i$ 
is $(\delta_i^{(1)}, \delta_i^{(2)}, \dots, \delta_i^{(T)})\,,$
which can be summarized by a score using the standard deviation of changes (SDC) \cite{martinez2017does} (to express its dispersion) given as follows
\begin{equation}
\label{eq:mac}
   \sigma_i = \sqrt{\frac{1}{T}\sum_{t=1}^T \left(\delta_i^{(t)}-\frac{1}{T}\sum_{t=1}^T \delta_i^{(t)}\right)^2} \,. 
\end{equation}
%We expect anomalous candidates (or problematic data points that have potential to hinder model learning ability) to have high variance in the dynamic space.
Let $N$ denote the size of the data set. Our dynamic space is one-dimension, given by $\Sigma=\left(\sigma_i\right)_{i=1}^{N}$.

Cosine similarity is a useful metric in anomaly detection due to its scale invariance, effectiveness with high-dimensional data, and interpretability, as it focuses on the angle between vectors rather than their magnitude. This property makes it robust to noise and computationally efficient, particularly for large data sets. However, it also has limitations, such as sensitivity to small variations, loss of information regarding magnitude, and a reliance on linear relationships, which may not adequately capture complex anomalies. Additionally, its effectiveness is contingent upon appropriate vector representation; poor feature selection can lead to misleading results.

We employ the \(\ell_2\) norm for training and cosine similarity to measure changes during the training process. Our choice to employ the \(\ell_2\) norm for training and cosine similarity for constructing a dynamic space where anomalies can be easily identified is grounded in our experimental findings. The \(\ell_2\) loss function is particularly effective for anomaly detection because it is sensitive to outliers, a well-documented characteristic in the literature (for example in \cite{goto2005robust}). In contrast, cosine similarity serves as a suitable metric for tracking changes in loss over time, providing insights into the direction of those changes. Each metric fulfills a distinct role; using the same metric for both training and dynamic space computation tends to yield suboptimal results.

%As observed, the proposed method relies heavily on two key components: autoencoders and isolation forests. Consequently, the quality of the dynamic space is inherently dependent on the effectiveness of the autoencoder. If the autoencoder fails to generate an embedding space that adequately separates anomalies from normal data points, some anomalies may remain undetected.
%We assume that the dynamic space generated by autoencoders will be of sufficient quality for our objective, as it has been reported in the literature that autoencoders are a state-of-the-art method for unsupervised anomaly detection.
\subsection{Standard deviation of changes (SDC) for dynamics scores}
We expect the dynamics scores at anomalies to have high dispersion. 
SDC is a good choice for measuring  data dispersion.  Calculating SDC of a sequence can provide valuable insights into its dynamics, particularly in understanding how the dispersion of the data evolves over time. There are three main reasons we choose SDC. First, it quantifies the spread of data around the mean; therefore, a changing SDC indicates variations in the data's variability over time. Second, since it is sensitive to anomalies, it reflects changes caused by  them, making it a useful indicator of data dynamics. Lastly, abrupt changes in SDC might signify shifts in the underlying  behavior.
Higher SDC values indicate greater dispersion within the sequence.  (But other metrics might be used.)

Considering dynamic score $\sigma_i$ associated to a data point $x_i$ (as in \eqref{eq:mac}), we assume the following:
\begin{itemize}
    \item Anomalies produce dynamics scores that deviate from those of inliers, $\mathbb{E}[\sigma_i | x_i \in X^+]\neq \mathbb{E}[\sigma_i | x_i \in X^-]$ where $\mathbb{E}[\sigma_i | x_i \in X]$ denotes the expected dynamics corresponding to data points in $X$. This allows segmentation of dynamics into two groups (inliers vs anomalies) \cite{iglesias2025anomaly}.
\item Over iterations, the reconstruction errors tend to converge to a stable level \cite{osada2023out}, $
e_i^{(t')} \approx e_i^{(t_0)}\,$ for $t'> t_0$.
So we set $T$ to be relatively small to produce distinctive dynamics.  Otherwise, all data points would correspond to dynamics  having almost the same distribution or scores.
\end{itemize}

 \begin{algorithm}
\begin{algorithmic}[1]
\Function{bopeto}{$X$}
\State $T, N \gets 10, \# X$
\State $N_\Delta = \frac{3N}{10}$
%\State $X_\Delta \gets \text{synthetic-data-generation}(X, %N_\Delta)$
%\State $X'\gets X\cup X_\Delta$
\State $h^{(0)}_\Theta \gets $ \text{auto-encoder(parameters)}
\For{$t \in \{1, 2, \dots, T\}$} 
\State $h^{(t)}_\Theta\gets $\text{update}($h^{(t-1)}_\Theta, X$)
\For{$x_i \in X$} 
\State $e^{(t)}_i \gets \text{cosine-similarity} \left(x_i, h^{(t)}_\Theta(x_i)\right)$ 
\EndFor
\EndFor
\State $\Sigma\gets \emptyset$
\For{$x_i \in X$} 
\State $\sigma_i\gets \text{dynamics-score}(e^{(1)}_i, \cdots, e^{(T)}_i)$
\State $\Sigma\text{.add(}\sigma_i)$
\EndFor
\State $(\hat{Y}) \gets \text{traditional-OoD}(\Sigma)$
\State \textbf{return} $\hat{Y}$
\EndFunction
\end{algorithmic}
\caption{Bopeto: data cleaning.}\label{algo:algo}
     
 \end{algorithm}

\subsection{Bopeto algorithm}
%We now describe the two main steps of Bopeto as indicated in Figure \ref{fig:archi}
\subsubsection{Dynamic space}
determination of dynamic space involves tracking of the cosine similarities per data point, and calculation of dynamics scores (a dynamics score associated to a data point is a value that represents its dynamics). Recall that the dynamic space has only one dimension. %Let $X_\Delta$ denote the synthetic data set. The new training data set is then $X'=X\cup X_\Delta$.
After  $T$ training iterations, the dynamics score of a data point $x_i \in X'$  is denoted by $\sigma_i$, which is obtained using  \eqref{eq:mac}. $\Sigma$ denotes the set of all dynamics scores. 
%plots
Our algorithm of Bopeto is given in Algorithm \ref{algo:algo} and its code could be found at \url{https://github.com/jmf-mas/jmf_bopeto}.
\subsubsection{Theoretical justification for separating unlabeled anomalies}
The core intuition behind our approach is that unlabeled anomalies, even when embedded in contaminated training data, exhibit irregular reconstruction behavior over time. By analyzing the dynamics of cosine similarity between input data points and their reconstructions during training, we expect to capture temporal fluctuations that are characteristically different for anomalies and inliers.

Formally, we assume:
$$
\mathbb{E}[\sigma_i \mid x_i \in X^+] \neq \mathbb{E}[\sigma_i \mid x_i \in X^-],
$$

where $\sigma_i$ represents the standard deviation of cosine similarities across training epochs. This deviation results from the poor and unstable reconstruction of anomalous data points by autoencoders. In contrast, inliers tend to produce stable similarity trajectories. The expected separation between these dynamics allows for effective segmentation of data in the one-dimensional dynamic space.

This observation is supported by recent work (e.g., \cite{iglesias2025anomaly}) showing that dispersion-based measures of anomaly scores offer stronger discriminant power than static values alone. 
\subsubsection{Theoretical guarantees for data cleaning performance}
Let $d(x)$ be a dynamics-based anomaly score (e.g., SDC) and let $T$ be a threshold chosen via a traditional unsupervised anomaly detection (e.g., isolation method, local outlier factor or one-class SVM). Let define

\[
\text{Precision} = \frac{|X^+ \cap X_{\text{removed}}|}{|X_{\text{removed}}|}, \quad
\text{Recall} = \frac{|X^+ \cap X_{\text{removed}}|}{|X^+|}.
\]

Assuming that anomalies occupy lower-density regions of the dynamic space and exhibit significantly higher dispersion, we can show that, under appropriate thresholding, the cleaning process maximizes contamination reduction while preserving a large fraction of inliers. In practice, our method should be useful only when it retains at least $90\%$ of inliers while removing up to $100\%$ of contamination.

%discussion

\subsubsection{Anomaly identification and treatment}

Our objective is to distinguish the dynamics scores of inliers from those of anomalies. 
We hypothesize that anomalous candidates  would disperse towards less densely populated regions of the dynamic space.  To substantiate this objective, we apply IF on the dynamic space to detect deviated dynamics scores. Let $\hat{Y}=\{\hat{y}_1, \hat{y}_2, \dots, \hat{y}_{N+N_\Delta}\}$ ($\hat{y}_i=0$ for normality and $\hat{y}_i=1$ for anomaly) denote the set of predictions provided by a traditional anomaly detection method respectively.  %We use Scikit-Learn \cite{pedregosa2011scikit} for IF.
%The \texttt{fit\_predict} and \texttt{decision\_function} methods of IF  produces negative measures for anomalies and positive measures for inliers, with its decision threshold set to zero. While the threshold of zero maximizes  anomaly detection in the dynamic space, it also increases the risk of incorrectly classifying some inliers. %To preserve most inliers while still detecting as many anomalies as possible, the threshold needs to be adjusted. We randomly adjust the threshold $\tau$ to be between the 60th and 70th percentiles of negative measures obtained from IF. Subsequently, we normalize these measures into the range [0, 1] to convert them into weights.

%Potential anomalies correspond to dynamics with relatively small weights $\omega_i$.
We achieve \textit{hard cleaning} where predicted anomalies are all excluded from the training data set (refers to the red line in Figure \ref{fig:distr2}).  % For soft cleaning, the loss of each data point will be factored by its weight ($\omega_i\|x_i-\hat{x}_i^{(t)}\|$ as in \eqref{eq:loss}) at each iteration $t$. 
%All synthetic data points are removed after cleaning.  
We will apply  both types of cleaning on deep models, while only hard cleaning on shallow  models.

\textbf{Complexity analysis}. Bopeto is a sequential combination of autoencoders and traditional unsupervised anomaly detection, so its time complexity is the sum of the complexities of both techniques. Recall that $T$ denotes the number of epochs, $N$ the number of training samples, and let $p$ be the number of data attributes, $h$ the number of hidden layers (for the autoencoders), and $n_t$ the number of iterations in the traditional unsupervised anomaly detection method . The overall time complexity of Bopeto is given by $O(TNph + n_t N \log(N))$.

\section{Anomaly detection investigation}
We now  assess the impact of  Bopeto  as a preprocessing anomaly detection approach to unsupervised anomaly detection (as a classification task).
\subsection{Experimental setup}
\subsubsection{Data sets used for experimental investigation}
We now introduce various benchmark data sets, as our intention is to discuss results at each step of the system: (1) data cleaning and, (2) unsupervised anomaly classification. We primarily focus on cybersecurity data sets, given the paramount importance of anomaly detection within it \cite{lim2024future}. However, we also include two other data sets from medical  and banking systems, and four image data sets, for broader context and comparison. %We use PyTorch \cite{paszke2019pytorch} for implementation.

 \begin{table}
	\centering
	\scalebox{.7}{
		\begin{tabular}{|c|c|c|c|c|}
			\hline
			\textbf{Data set} & \textbf{\# samples} & \textbf{\# attributes} & \textbf{untreated anomaly (\%)}\\
			\hline
         	CICIOT  & $416985$ &$41$ & $45$\\
            \hline
   			CIFAR-10 & $6000$ &$1024$ & $90$\\
			\hline
         	CREDIT & $234333$ &$30$ & $0.2$\\
			\hline
            ECG  & $4998$ &$140$ & $58$\\
			\hline
      		IDS & $430256$ &$95$ & $43$\\
            %\hline
			%NSL & $148517$ &$125$&$48$ \\
			\hline
   			KITSUNE & $210171$ &$116$ & $23$\\
			\hline
			MNIST & $7000$ &$784$&$90$ \\
                                    \hline
            MVTec  & $5354$ &$1048576$ & $58$\\
			\hline
      		Visa & $10821$ &$95$ & $11$\\

			\hline
	\end{tabular}}
	\caption{The characteristics of data sets. %Recall that training data contains normal datum only.
 }
	\label{tab:dataset}
\end{table}
\begin{figure*}[h]
\centering
\includegraphics[width=14cm]{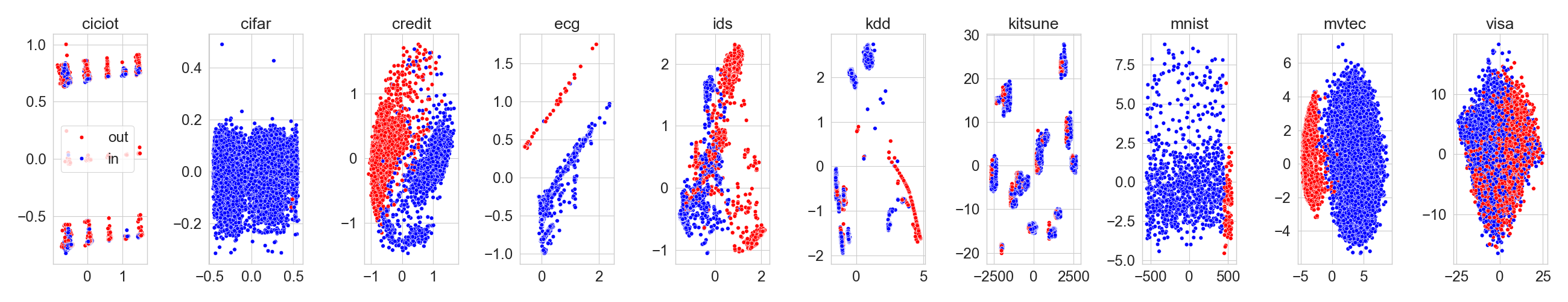}
\caption{Visualisation of training data sets.}
\label{fig:training}
\end{figure*}

\begin{figure*}[!ht]
    \centering
        \includegraphics[width=\textwidth]{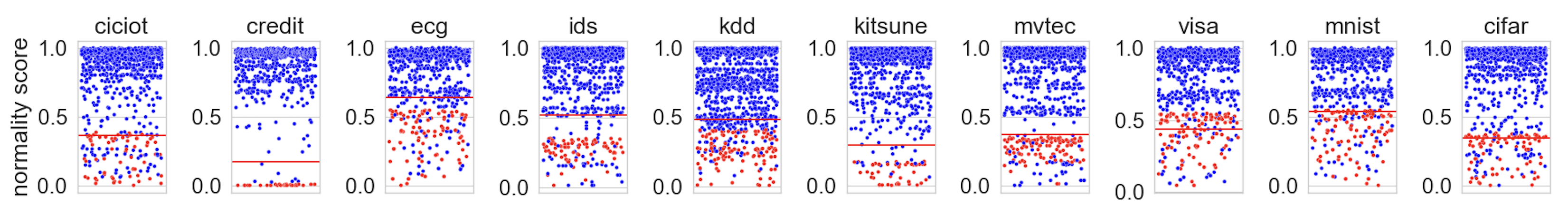}  
    \caption{A sample of the dynamic space. Blue points are inliers and red are anomalies. A t‑SNE projection is used for visualization.}
    \label{fig:dynamics_one}
\end{figure*}
We use ten  data sets for experiments, whose distributions are visualized in Figure \ref{fig:training}, and 
characteristics  are summarized in Table \ref{tab:dataset}:
(1) CICIOT: Real-time IoT network data containing 33 executed attacks \cite{neto2023ciciot2023}.  
(2) CIFAR‑10: Color images from 10 classes (airplane, automobile, bird, cat, deer, dog, frog, horse, rabbit, ship) \cite{krizhevsky2009learning}.  
(3) CREDIT: Financial transaction records labeled as fraudulent or legitimate \cite{warghade2020credit}.  
(4) ECG: Heart electrical activity recordings capturing normal and anomalous waveforms \cite{khan2021ecg}.  
(5) IDS (revised): Simulated complex network traffic with multiple attack types \cite{sharafaldin2018toward}.  
(6) KDD (revised): Intrusion detection data with simulated military traffic and diverse attacks \cite{tavallaee2009detailed}.  
(7) KITSUNE: Nine network-attack data sets from an IP‑based surveillance system or IoT network \cite{mirsky2018kitsune}.  
(8) MNIST: Grayscale handwritten digits (0–9) \cite{lecun2010mnist}. MNIST and CIFAR-10 are adapted for anomaly detection by treating one class as inliers.  
(9) MVTec: Industrial anomaly-detection benchmark with 15 object/texture categories, defect‑free training images, and defective/non‑defective test images \cite{bergmann2019mvtec}.  
(10) Visa: Anomaly-detection data set with objects from 12 classes across 3 domains \cite{zou2022spot}.

For clarity and consistency, we will start discussing the first six data sets, which are tabular, and then we will include all data sets, including the last four, which are image data sets.

\subsubsection{Data contamination}
We  begin by creating contaminated  data sets with differing ratios of contamination. Bopeto will be used to minimize contamination as much as possible.
For the  cleaning process, we will use 60\% of inliers for training while reserving 30\% of anomalies for contamination. We  sample instances from the contamination set to corrupt the training data set. 
To sample the contamination set, we partition it into $9$ disjoint subsets. We progressively incorporate each of the first $n$ subsets into the training data set (we begin with the first subset, followed by the first two subsets, and so on).
The all remaining 40\% of inliers and randomly selected fractions of the remaining 70\% of anomalies will be retained for testing various anomaly detection models for classification task. All data sets are normalized into the range $[0, 1]$ using the minmax function. 

\subsubsection{Models used}
We consider AE, isolation forest (IF), local outlier factor (LOF), SVDD \cite{ruff2018deep}, DAGMM \cite{zong2018deep}, ALAD \cite{zenati2018adversarially} and DSEBM \cite{zhai2016deep} for unsupervised anomaly detection.  We also apply IAD  in AE and SVDD as in \cite{kim2023iterative}. We employ two ways to evaluate IAD against Bopeto. Firstly, we assess the influence of both Bopeto and IAD in AE and SVDD individually. Secondly, we evaluate the impact of Bopeto in IAD. NeutrALAD \cite{qiu2022latent}  and DUAD \cite{li2021deep}, discussed earlier, and  OCSVM \cite{alam2020one} are not discussed intensively in this analysis since they require information on the contamination ratio, which is assumed unavailable. IF and LOF also need information on contamination ratio, but if not explicitly provided, they estimate its value internally. IF, LOF, AE, SVDD are  standard models, while others are advanced. 
We allocate 20\% of the test data as a validation set for threshold determination and early stop verification. We run each scenario of Bopeto 5 times to generate error bars.
The hyperparameters for models are as follows: a weight decay of $10^{-6}$, batch sizes of 64, $100$ epochs, patience of 5 for early stop, learning rates will be set to $10^{-3}$ and Adam as optimizer. We consider 10 rounds and 10 epochs per round for IAD to make a total of 100 epochs (as set in the original paper \cite{kim2023iterative}).  To rigorously investigate performance robustness, all models begin with identical initial parameters.  We report F$_1$ score, accuracy, precision and recall to measure classification quality. %We use the original published code of all baselines. 
In what follows, Bopeto data will refer to ''cleaned'' data. We use the $p$-value and effect size \cite{effectsize} to measure differences between performances on the original and Bopeto data.
%The code is available at \url{https://github.com/jmf-mas/bopeto} and \url{https://github.com/djeffkanda/robustad_with_noise}.

%\subsection{Performance degradation} 
%\begin{figure}[h]
%       \centering
%        \includegraphics[width=\textwidth]{degration_f1.pdf}
%\caption{Impact of contamination in performances of models using F$_1$ score.}
%\label{fig:degradation}
%\end{figure}
%\textbf{Results}. Model's performance degrades to data contamination, as illustrated in Figure \ref{fig:degradation}. Many research has shown the degradation of even advanced anomaly detection techniques when faced contaminated data. This underscores the importance of developing  anomaly detection models that can be less sensitive to the inaccuracy of the data set used during its training.
%Performance degradation is also observed with other performance metrics (precision, recall and accuracy).

Excluding data normalization, splitting, contamination, and some other preprocessing operations, the average training duration per contaminated data set for Bopeto was about 10 minutes, and for unsupervised anomaly detection was about 40 minutes.  Inference time was about 30 seconds per data set. There are 6 data sets, each with 9 contaminated versions, making a total of 60 training data sets.
Models were trained on a high-performance computer equipped with a multi-core 11th Gen Intel Core i7-11800H (2.30GHz) x 16, 32 GB of RAM, and an NVIDIA TGL GT1 using Ubuntu 20.04.

%While we may not have included all state-of-the-art unsupervised anomaly detection models, the proposed method is a preprocessing approach designed to clean datasets before they are used to train anomaly detection models. We anticipate that it will positively impact larger anomaly detection techniques.
\subsection{On quality of the data cleaning process}
Figure \ref{fig:dynamics_one} illustrates dynamic spaces. Most low-dense dynamics scores correspond to anomalies. Inliers with corresponding low-dense dynamics scores could either be problematic (OoD) or indicative of ineffectiveness in our approach in some cases. In some data (e.g., IDS), our approach faces challenges in distinguishing between inliers and anomalies. With CREDIT, the substantial variance of anomalies can be attributed to its low level of contamination. However, with other data sets, anomalies are observed to exhibit specific means and high variance in the dynamic space.  Although Bopeto does not always hold, but it indicates a better separation between anomalies and inliers  in most cases (compared to the original spaces as in Figure \ref{fig:training}).

\begin{figure}
    \centering
\includegraphics[width=9cm]{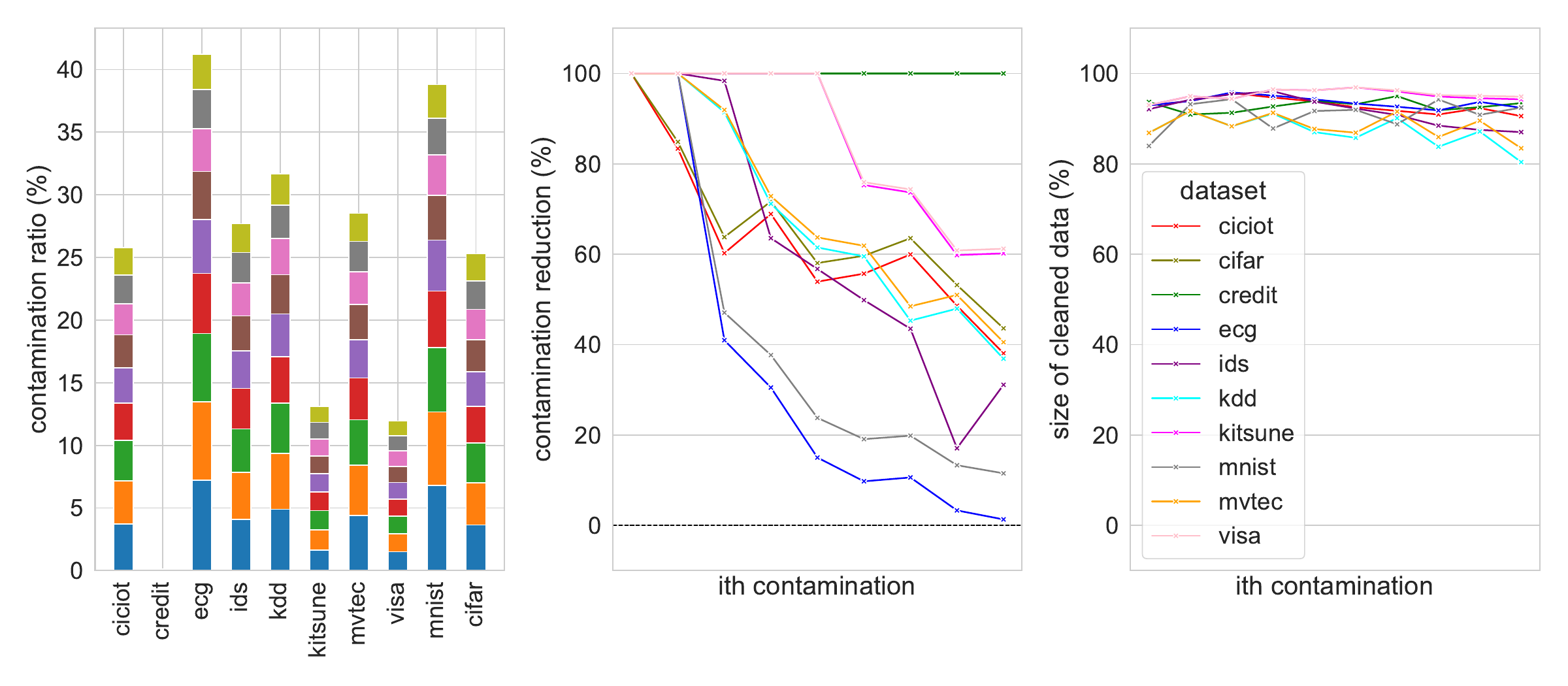}
\caption{(Left) shows the level of contamination per round across 9 different contamination ratios for each dataset. (Middle) displays the reduction in contamination ratio following data cleaning. (Right) demonstrates that Bopeto not only removes contamination but also retains at least 90\% of inliers.}
\label{fig:dynamics_contamination}
\end{figure}

Figure \ref{fig:dynamics_contamination} (left-hand side) illustrates the level of contamination per round (we consider 9 ratios of contamination) in each data set. For example, the contamination levels in ECG begin at 7.2\% and peak at 41.2\%. Figure \ref{fig:dynamics_contamination} (middle) illustrates the contamination ratio reduction after data cleaning, demonstrating that significantly less contamination remains post-cleanup. The ``cleaned`` data set can subsequently be employed by various unsupervised anomaly detection models.
 But, reducing data contamination is deemed acceptable only if inliers from the original data set are preserved. Figure \ref{fig:dynamics_contamination} (right-hand side) reveals that Bopeto not only eliminates contamination but also retains most inliers (at least $90\%$) from the original data.  However, one downside of Bopeto is that it identifies ``anomalies`` even from an uncontaminated data set. Since no information on the contamination ratio is provided, Bopeto finds data points whose corresponding dynamics seem anomalous. 
 This could pose a problem depending on how informative those data points could be. If they are faithful inliers, then our approach may lead to underperformance. However, if a model trained on the Bopeto version of uncontaminated data outperforms its own performance when trained on the original data, it may suggest that the discarded inliers were problematic. This is important as, in reality, uncontaminated training data set might still contain some problematic data points, making models struggle to learn a more informative data representation. Further human-involved investigation  on problematic data points might be required for data validation.
Also, Figure \ref{fig:dynamics_contamination} shows that applying Bopeto to a clean dataset leads to consistently strong performance. When the underlying data contains little to no contamination, Bopeto aligns well with the true structure of the normal class, allowing downstream anomaly‑detection models to benefit from richer and more representative training data. 

\begin{figure*}[!ht]
    \centering
        \includegraphics[width=0.9\textwidth]{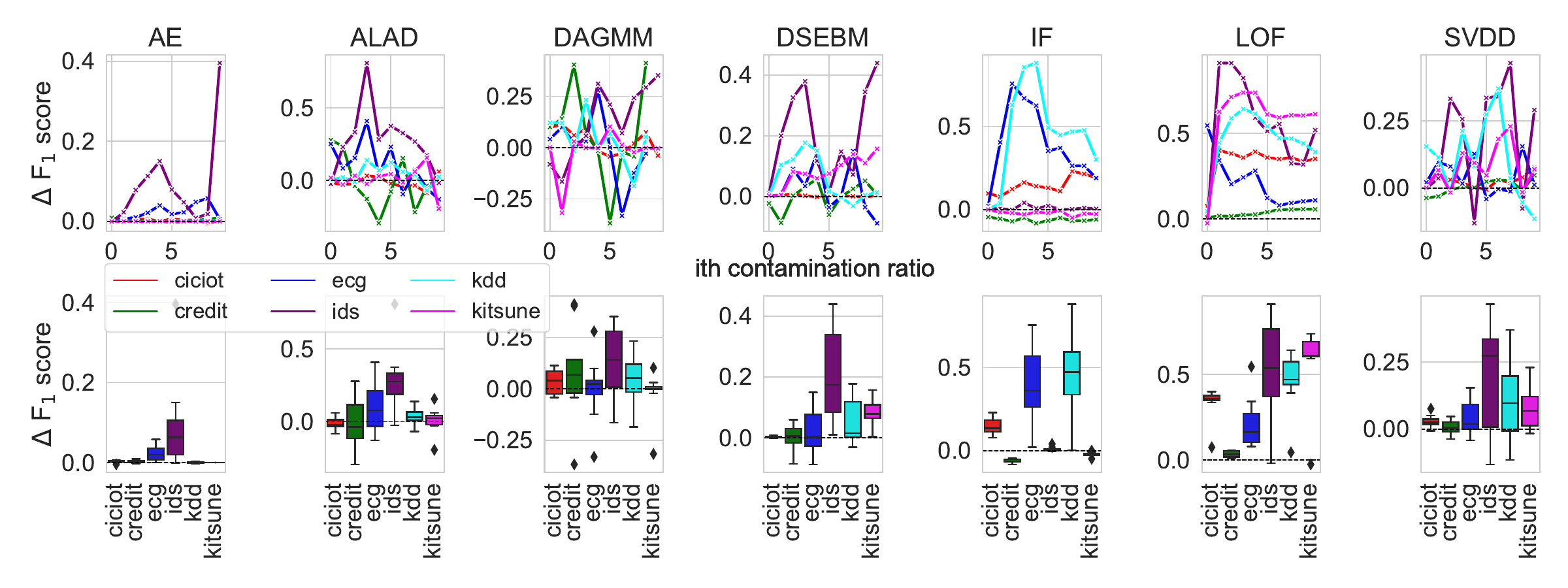}
    \caption{Differences in F$_1$ scores between models trained on Bopeto data and those trained on the original data are shown. Positive values indicate better performance with Bopeto. For instance, if the F$_1$ scores are 0.7 (Bopeto) and 0.55 (original), then $\Delta \text{F}_1$ = +0.15. The first row reports detailed $\Delta \text{F}_1$ values for each contamination ratio (as in Figure \ref{fig:dynamics_contamination}), while the second row shows the distributions of $\Delta \text{F}_1$ scores across contamination ratios.}
    \label{fig:robustness}
\end{figure*}

\begin{figure*}[!ht]
    \centering
        \begin{subfigure}[b]{1\textwidth}
        \includegraphics[width=\textwidth]{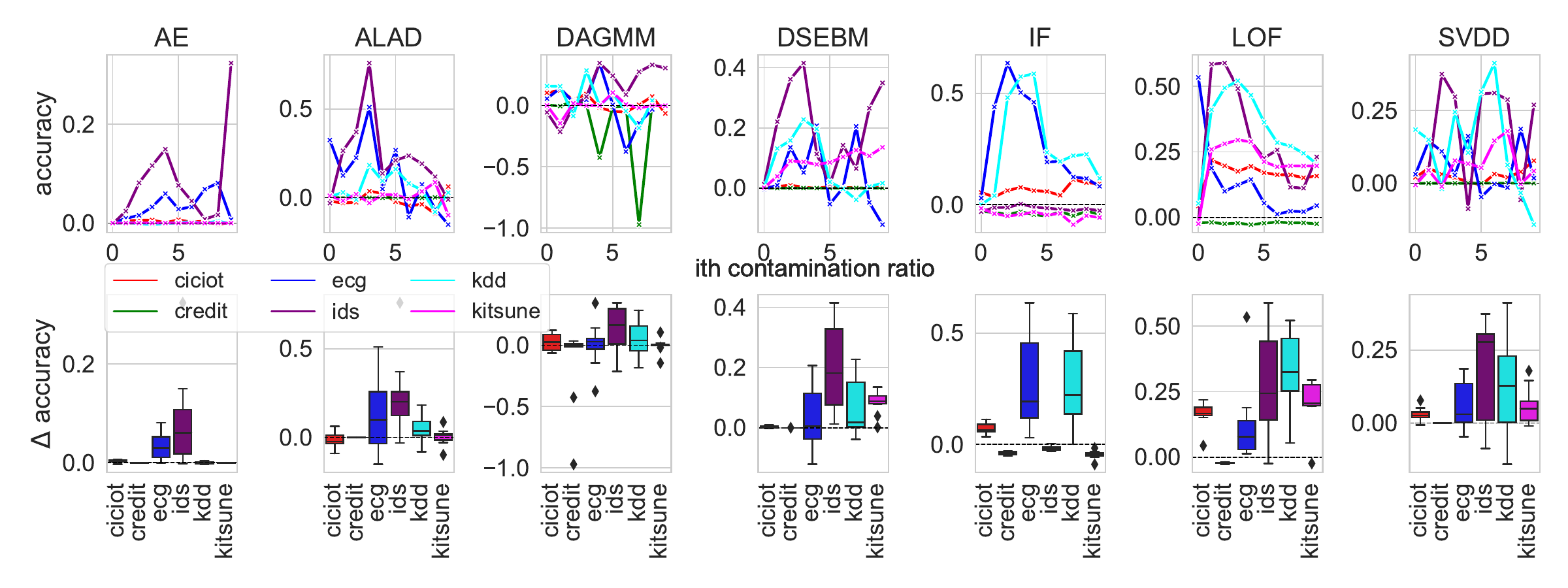}
        \caption{}
        \label{fig:accuracy_boxplot_bop_difference}
    \end{subfigure}
      \begin{subfigure}[b]{1\textwidth}
        \includegraphics[width=\textwidth]{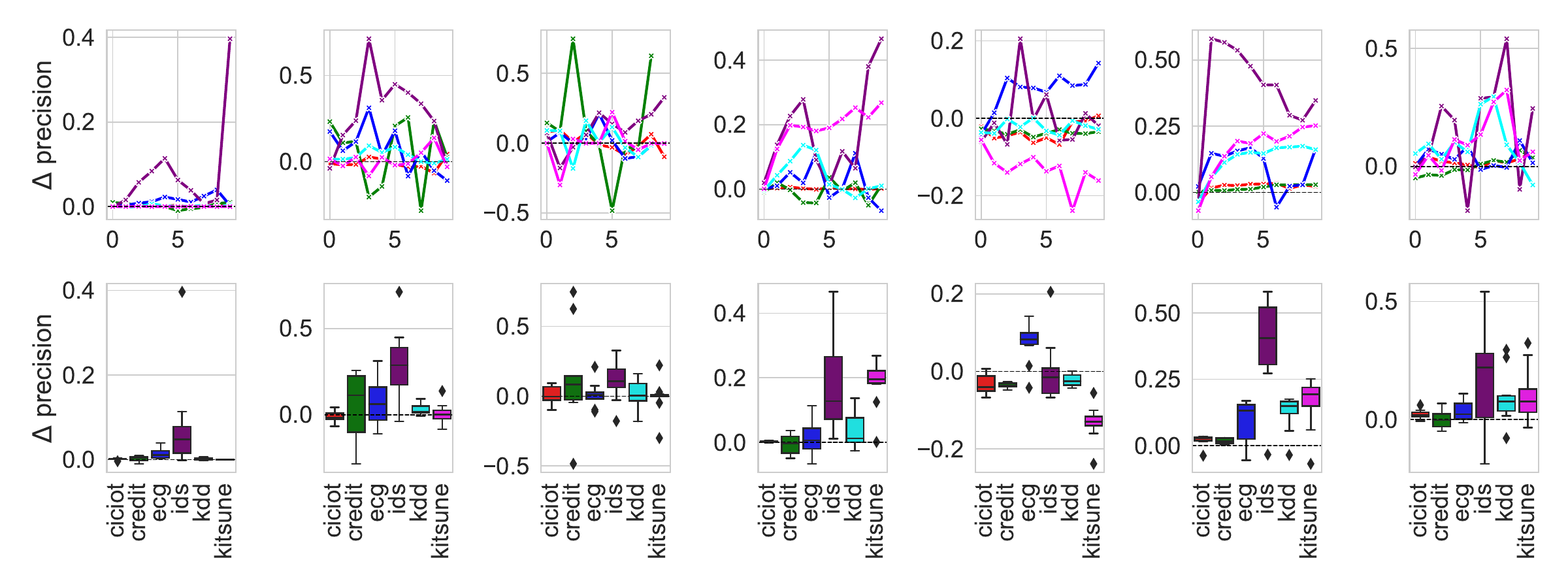}
        \caption{}
        \label{fig:precision_boxplot_bop_difference}
    \end{subfigure}
    \begin{subfigure}[b]{1\textwidth}
        \includegraphics[width=\textwidth]{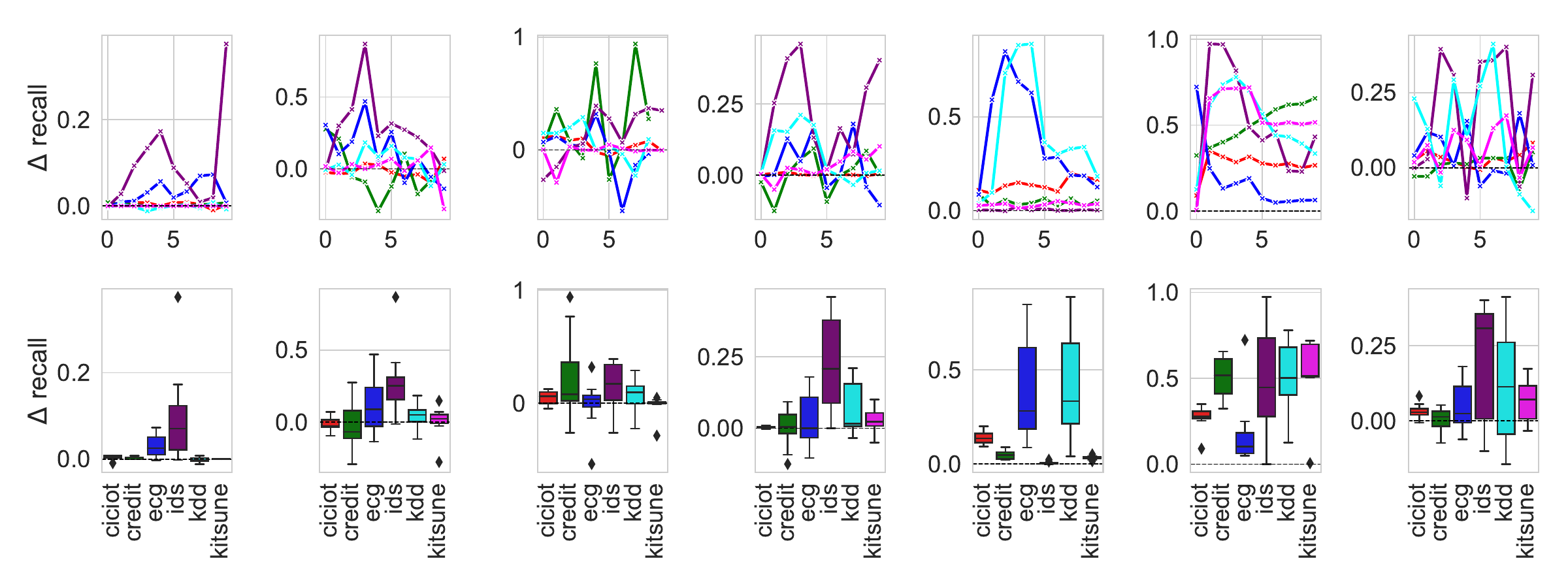}
        \caption{}
        \label{fig:recall_boxplot_bop_difference}
    \end{subfigure}
    \caption{
    Differences between models trained on Bopeto data and  the original data using (\ref{fig:accuracy_boxplot_bop_difference}) accuracy, (\ref{fig:precision_boxplot_bop_difference}) precision and (\ref{fig:recall_boxplot_bop_difference}) recall. Positive values indicate the superiority of Bopeto. First row represents detailed performance differences per ith contamination ratio and second row represents their distributions across contamination ratios.}
    \label{fig:boxplot_bop_difference_all}
\end{figure*}

\begin{table*}[h]
	\centering
\scalebox{0.6}{\begin{tabular}{|c|c|c|c|c|c|c|c|c|c|c|c|}
\hline
\multicolumn{2}{|c|}{} & \textbf{CICIOT} & \textbf{CIFAR} & \textbf{CREDIT} & \textbf{ECG} & \textbf{IDS} & \textbf{KDD} & \textbf{KITSUNE} & \textbf{MNIST} & \textbf{MVTEC} & \textbf{VISA}   \\
\hline
\multirow{2}{*}{AE}& original & 0.73 $\pm$ 0.00 & 0.74 $\pm$ 0.10  & 0.47 $\pm$ 0.00 & 0.90 $\pm$ 0.05 & 0.60 $\pm$ 0.16 & 0.93 $\pm$ 0.00 &   0.56 $\pm$ 0.00 & 0.80 $\pm$ 0.55 & 0.76 $\pm$ 0.40 & 0.75 $\pm$ 0.35   \\ 
                   & Bopeto & 0.73 $\pm$ 0.00 & \textbf{0.79 $\pm$ 0.15} & 0.47 $\pm$ 0.00 & \textbf{0.92 $\pm$ 0.05} & \textbf{0.70 $\pm$ 0.09} & 0.93 $\pm$ 0.00 &   0.56 $\pm$ 0.00 & \textbf{0.82 $\pm$ 0.05} & \textbf{0.77 $\pm$ 0.25} & 0.75 $\pm$ 0.45\\
\hline\hline
\multirow{2}{*}{SVDD}& original & 0.61 $\pm$ 0.03 & 0.66 $\pm$ 0.13 & 0.66 $\pm$ 0.03 & 0.76 $\pm$ 0.09 &   0.50 $\pm$ 0.33 & 0.72 $\pm$ 0.14 & 0.60 $\pm$ 0.09 & 0.65 $\pm$ 0.20 & 0.66 $\pm$ 0.15 & 0.67 $\pm$ 0.12\\ 
                   & Bopeto &\textbf{0.64 $\pm$ 0.04} &\textbf{0.67 $\pm$ 0.14} & \textbf{0.67 $\pm$ 0.01} &  \textbf{0.81 $\pm$ 0.09} & \textbf{0.68 $\pm$ 0.36} & \textbf{0.83 $\pm$ 0.10} & \textbf{0.68 $\pm$ 0.04} &\textbf{0.68 $\pm$ 0.20} &0.66 $\pm$ 0.05 &\textbf{0.69 $\pm$ 0.20}\\
\hline \hline
\multirow{2}{*}{IF}& original & 0.39 $\pm$ 0.08 & \textbf{0.15 $\pm$ 0.01} & 0.28 $\pm$ 0.27 & 0.01 $\pm$ 0.00 & 0.24 $\pm$ 0.36 & \textbf{0.58 $\pm$ 0.01} & 0.44 $\pm$ 0.05 & 0.42 $\pm$ 0.06 & 0.45 $\pm$ 0.09 & 0.48 $\pm$ 0.08\\ 
                   & Bopeto & \textbf{0.53 $\pm$ 0.04} & 0.08 $\pm$ 0.01 &  \textbf{0.67 $\pm$ 0.27} & \textbf{0.02 $\pm$ 0.01}& \textbf{0.70 $\pm$ 0.24} & 0.56 $\pm$ 0.01 & \textbf{0.50 $\pm$ 0.06} & \textbf{0.53 $\pm$ 0.07} & \textbf{0.49 $\pm$ 0.05} & \textbf{0.55 $\pm$ 0.04}\\
\hline \hline
\multirow{2}{*}{LOF}& original & 0.22 $\pm$ 0.13 & 0.06 $\pm$ 0.02 & 0.13 $\pm$ 0.11 & 0.16 $\pm$ 0.29 &  0.33 $\pm$ 0.22 & 0.18 $\pm$ 0.24 & 0.30 $\pm$ 0.07 & 0.27 $\pm$ 0.06 & 0.35 $\pm$ 0.10 & 0.29 $\pm$ 0.07\\ 
                   & Bopeto &\textbf{0.56 $\pm$ 0.06} & \textbf{0.10 $\pm$ 0.00} & \textbf{0.34 $\pm$ 0.25} & \textbf{0.70 $\pm$ 0.22} & \textbf{0.80 $\pm$ 0.15} & \textbf{0.76 $\pm$ 0.08} & \textbf{0.42 $\pm$ 0.06} & \textbf{0.40 $\pm$ 0.05} & \textbf{0.48 $\pm$ 0.07} & \textbf{0.41 $\pm$ 0.06}\\
\hline \hline
\multirow{2}{*}{ALAD}& original & \textbf{0.72 $\pm$ 0.03} &  \textbf{0.29 $\pm$ 0.12} & 0.61 $\pm$ 0.13 & 0.29 $\pm$ 0.36 & 0.88 $\pm$ 0.05 & 0.60 $\pm$ 0.08 & 0.49 $\pm$ 0.05 & 0.52 $\pm$ 0.07 & 0.66 $\pm$ 0.08 & 0.63 $\pm$ 0.09\\
                   & Bopeto & 0.71 $\pm$ 0.03 &  0.28 $\pm$ 0.11 & \textbf{0.71 $\pm$ 0.19} & \textbf{0.56 $\pm$ 0.37} & \textbf{0.92 $\pm$ 0.05} & \textbf{0.61 $\pm$ 0.03} & \textbf{0.55 $\pm$ 0.06} & \textbf{0.58 $\pm$ 0.05} & \textbf{0.68 $\pm$ 0.06} & \textbf{0.66 $\pm$ 0.05}\\
\hline \hline
\multirow{2}{*}{DAGMM}& original & 0.64 $\pm$ 0.06 & 0.12 $\pm$ 0.12 & 0.76 $\pm$ 0.11 &  0.78 $\pm$ 0.20 & 0.78 $\pm$ 0.06 & \textbf{0.55 $\pm$ 0.03} & 0.50 $\pm$ 0.07 & 0.54 $\pm$ 0.09 & 0.65 $\pm$ 0.10 & 0.63 $\pm$ 0.09\\ 
                   & Bopeto & \textbf{0.67 $\pm$ 0.05} & \textbf{0.20 $\pm$ 0.16} & \textbf{0.77 $\pm$ 0.14} & \textbf{0.91 $\pm$ 0.12} & \textbf{0.82 $\pm$ 0.10} & 0.53 $\pm$ 0.10 & \textbf{0.55 $\pm$ 0.05} & \textbf{0.61 $\pm$ 0.06} & \textbf{0.69 $\pm$ 0.08} & \textbf{0.68 $\pm$ 0.07}\\
\hline \hline
\multirow{2}{*}{DESBM}& original & 0.74 $\pm$ 0.04 &  0.48 $\pm$ 0.03 & 0.86 $\pm$ 0.08 &  0.55 $\pm$ 0.25 & 0.81 $\pm$ 0.06 & 0.50 $\pm$ 0.05 & 0.62 $\pm$ 0.06 & 0.63 $\pm$ 0.07 & 0.71 $\pm$ 0.09 & 0.70 $\pm$ 0.07\\ 
                   & Bopeto & 0.74 $\pm$ 0.00 & 0.48 $\pm$ 0.02 & \textbf{0.89 $\pm$ 0.08} & \textbf{0.75 $\pm$ 0.20} & \textbf{0.86 $\pm$ 0.09} & \textbf{0.58 $\pm$ 0.01} & \textbf{0.66 $\pm$ 0.06} & \textbf{0.68 $\pm$ 0.05} & \textbf{0.73 $\pm$ 0.07} & \textbf{0.72 $\pm$ 0.06}\\
\hline
\hline
\multicolumn{2}{|c|}{Improvement ratio (\%)} &  \textbf{85.7} & \textbf{71.4}  & \textbf{100} & \textbf{100} &  \textbf{100} & \textbf{85.7} & \textbf{100} & \textbf{100} & \textbf{71.4} & \textbf{87.5}\\
\hline
\end{tabular}}
	\caption{Different rows (except the last row) represent average F$_1$-scores. Last row counts successful cases where Bopeto shows superiority for anomaly detection. Improvement ratio represents the ratio of contamination identified by Bopeto, e.g, if there are 1,000 malicious instances in the training data set and the model identifies 700 of them, the ratio would be 70\%.
 }
	\label{tab:average}
\end{table*}

\begin{table*}[h]
    \centering
    \scalebox{0.6}{\begin{tabular}{|l|c|c|c|c|c|c|c|c|c|c|c|c|}
        \hline
        & \textbf{CICIOT} & \textbf{CIFAR} & \textbf{CREDIT} & \textbf{ECG} & \textbf{IDS} & \textbf{KDD} & \textbf{KITSUNE} & \textbf{MNIST} & \textbf{MVTEC} & \textbf{VISA} & \textbf{p-value} & \textbf{effect size} \\
        \hline
        \multicolumn{13}{|c|}{Differences of F$_1$ scores}\\
        \hline
\textbf{AE} & \textbf{0.00} & \textbf{0.10} & \textbf{0.00} & \textbf{0.02} & \textbf{0.10} & \textbf{0.00} & \textbf{0.00} & \textbf{0.06} & \textbf{0.04} & \textbf{0.04} & 3.32$\times 10^{-13}$ & \textbf{0.45} \\ \hline
\textbf{SVDD} & \textbf{0.03} & \textbf{0.22} & \textbf{0.01} & \textbf{0.05} & \textbf{0.18} & \textbf{0.11} & \textbf{0.08} & \textbf{0.20} & \textbf{0.16} & \textbf{0.10} & 4.21$\times 10^{-11}$ & \textbf{0.49} \\ \hline
\textbf{IF} & \textbf{0.14} & \textbf{0.02} & \textcolor{red}{-0.07} & \textbf{0.39} & \textbf{0.01} & \textbf{0.46} & \textcolor{red}{-0.02} & \textbf{0.12} & \textbf{0.12} & \textbf{0.11} & 1.41$\times 10^{-15}$ & \textbf{0.54} \\ \hline
\textbf{LOF} & \textbf{0.34} & \textbf{0.21} & \textbf{0.04} & \textbf{0.21} & \textbf{0.54} & \textbf{0.47} & \textbf{0.58} & \textbf{0.06} & \textbf{0.09} & \textbf{0.07} & 5.53$\times 10^{-23}$ & \textbf{0.61} \\ \hline
\textbf{ALAD} & \textcolor{red}{-0.01} & \textbf{0.12} & \textcolor{red}{-0.01} & \textbf{0.10} & \textbf{0.27} & \textbf{0.04} & \textbf{0.01} & \textbf{0.04} & \textbf{0.10} & \textbf{0.03} & 1.41$\times 10^{-23}$ & \textbf{0.55} \\ \hline
\textbf{DAGMM} & \textbf{0.03} & \textbf{0.12} & \textbf{0.08} & \textcolor{red}{-0.01} & \textbf{0.13} & \textbf{0.04} & \textcolor{red}{-0.02} & \textbf{0.09} & \textbf{0.10} & \textbf{0.02} & 6.32$\times 10^{-24}$ & \textbf{0.62} \\ \hline
\textbf{DESBM} & \textbf{0.00} & \textbf{0.15} & \textbf{0.00} & \textbf{0.03} & \textbf{0.20} & \textbf{0.05} & \textbf{0.08} & \textbf{0.09} & \textbf{0.05} & \textbf{0.09} & 8.55$\times 10^{-22}$ & \textbf{0.61} \\ \hline
\hline
\textbf{OCSVM} & \textbf{0.09} & \textbf{0.00} & \textbf{0.05} & \textbf{0.05} & \textbf{0.21} & \textbf{0.02} & \textbf{0.03} & \textbf{0.07} & \textbf{0.08} & \textbf{0.08} & 8.42$\times 10^{-09}$ & \textbf{0.65} \\ \hline
\textbf{DUAD} & \textbf{0.12} & \textbf{0.05} & \textbf{0.11} & \textbf{0.08} & \textbf{0.00} & \textcolor{red}{-0.04} & \textbf{0.06} & \textbf{0.13} & \textbf{0.12} & \textbf{0.22} & 7.65$\times 10^{-09}$ & \textbf{0.41} \\ \hline
\textbf{NeutrALAD} & \textbf{0.16} & \textbf{0.06} & \textbf{0.13} & \textbf{0.07} & \textbf{0.07} & \textbf{0.01} & \textbf{0.13} & \textbf{0.04} & \textbf{0.09} & \textbf{0.11} & 6.62$\times 10^{-25}$ & \textbf{0.45} \\ \hline
\multicolumn{13}{|c|}{Differences of AUC}\\
        \hline
\textbf{AE} & \textcolor{red}{-0.01} & \textbf{0.11} & \textbf{0.00} & \textbf{0.04} & \textbf{0.10} & \textcolor{red}{-0.02} & \textbf{0.01} & \textbf{0.08} & \textbf{0.07} & \textbf{0.05} & 1.15$\times 10^{-11}$ & \textbf{0.48} \\ \hline
\textbf{SVDD} & \textbf{0.05} & \textbf{0.19} & \textcolor{red}{-0.01} & \textbf{0.07} & \textbf{0.20} & \textbf{0.12} & \textbf{0.08} & \textbf{0.21} & \textbf{0.14} & \textbf{0.11} & 3.65$\times 10^{-10}$ & \textbf{0.51} \\ \hline
\textbf{IF} & \textbf{0.14} & \textbf{0.03} & \textcolor{red}{-0.05} & \textbf{0.36} & \textbf{0.02} & \textbf{0.43} & \textbf{0.02} & \textbf{0.01} & \textbf{0.13} & \textbf{0.11} & 1.05$\times 10^{-12}$ & \textbf{0.54} \\ \hline
\textbf{LOF} & \textbf{0.33} & \textbf{0.24} & \textbf{0.06} & \textbf{0.23} & \textbf{0.51} & \textbf{0.49} & \textbf{0.58} & \textbf{0.02} & \textbf{0.08} & \textbf{0.09} & 5.13$\times 10^{-21}$ & \textbf{0.62} \\ \hline
\textbf{ALAD} & \textcolor{red}{-0.01} & \textbf{0.15} & \textcolor{red}{-0.02} & \textbf{0.08} & \textbf{0.31} & \textbf{0.04} & \textcolor{red}{-0.01} & \textbf{0.04} & \textbf{0.10} & \textbf{0.02} & 9.72$\times 10^{-22}$ & \textbf{0.57} \\ \hline
\textbf{DAGMM} & \textbf{0.04} & \textbf{0.13} & \textbf{0.06} & \textcolor{red}{-0.02} & \textbf{0.15} & \textbf{0.07} & \textbf{0.03} & \textbf{0.01} & \textbf{0.12} & \textbf{0.02} & 5.98$\times 10^{-23}$ & \textbf{0.60} \\ \hline
\textbf{DESBM} & \textcolor{red}{-0.01} & \textbf{0.18} & \textbf{0.01} & \textbf{0.03} & \textbf{0.23} & \textbf{0.07} & \textbf{0.06} & \textbf{0.12} & \textbf{0.06} & \textbf{0.09} & 6.92$\times 10^{-20}$ & \textbf{0.62} \\ \hline
\hline
\textbf{OCSVM} & \textbf{0.11} & \textbf{0.02} & \textbf{0.05} & \textbf{0.07} & \textbf{0.22} & \textbf{0.02} & \textbf{0.01} & \textbf{0.01} & \textbf{0.09} & \textbf{0.07} & 8.01$\times 10^{-08}$ & \textbf{0.67} \\ \hline
\textbf{DUAD} & \textbf{0.12} & \textbf{0.07} & \textbf{0.10} & \textbf{0.10} & \textbf{0.02} & \textcolor{red}{-0.01} & \textbf{0.05} & \textbf{0.17} & \textbf{0.12} & \textbf{0.20} & 5.84$\times 10^{-08}$ & \textbf{0.43} \\ \hline
\textbf{NeutrALAD} & \textbf{0.16} & \textbf{0.10} & \textbf{0.13} & \textbf{0.07} & \textbf{0.04} & \textbf{0.01} & \textbf{0.12} & \textbf{0.03} & \textbf{0.09} & \textbf{0.09} & 7.22$\times 10^{-23}$ & \textbf{0.45} \\ \hline
\multicolumn{13}{|c|}{Reduction of statistical contamination}\\
\hline
\textbf{p-value} & 0.01 & 0.08 & 2.12$\times 10^{-23}$ & 0.09 & 0.08 & 0.00 & 1.13 & 0.21 & 0.12 & 1.19 & - & - \\ \hline
\textbf{effect size} & 0.22 & 0.30 & 0.31 & 0.33 &  & 0.40 & 0.21 & 0.22 & 0.01 & 0.10 & - & - \\
\hline
\end{tabular}}
    \caption{Differences of F$_1$ scores and AUC  of various unsupervised anomaly detectors on original and Bopeto data. Positive values indicate the positive impact of Bopeto on the underline baseline anomaly detection method. The last three methods are those relying on contamination ratio information. The last two columns are p-value and effect size between performances of methods on original and Bopeto data. Reduction of statistical contamination using density-based clustering and isolation forest for identification of anomalies on dynamic spaces.
The threshold for the p-value is set at 0.01. %Both the effect size and p-value indicate that density-based clustering and the isolation forest method produce similar results.
}
    \label{tab:f1}
\end{table*}

%Intuitively, performance should degrade proportionally to the contamination ratio. However, we noticed that performance degradation does not follow a downward trend of contamination ratios. An explanation could lie in the varied and unconventional distribution of anomalies. Certain anomalies might pose greater concerns than others, regardless of the contamination ratio. 

\subsection{Bopeto for robustness} 
As highlighted in Figure \ref{fig:robustness}, it becomes evident that various anomaly detection models trained on Bopeto data consistently outperform their counterparts trained on the original data. Bopeto effectively reduces a substantial ratio of contamination in the original data. This disparity underscores the substantial impact of data quality on anomaly detection.
In essence, Bopeto offers a more faithful representation of the underlying inliers data, thereby facilitating a more effective learning process. Consequently, even classic models produce competitive results when trained on Bopeto data. Further, contamination has a greater impact on some classic models than on advanced models. This explains why Bopeto has a more pronounced effect on standard models than on deep models.

However, with DAGMM for instance, Bopeto data could lead to underperformance. DAGMM varies significantly, thus complicating the determination of which data set produces better results. Performance variability can be attributed to its probabilistic nature.  
Also Bopeto is not effective when data is highly contaminated. This observation might suggest that when anomalies are overpopulated, our approach encounters difficulty in discerning their respective  dynamics. This phenomenon can be attributed to the expected rarity of anomalies, thus its overpopulation impacts the effectiveness of our method. 

Table \ref{tab:average} illustrates the average performance of models, highlighting the improvement in model performance achieved through Bopeto.
 We obtained a $p$-value of $5.95\times 10^{-27}$ and an effect size of $0.41$, which  indicate a significant positive impact of Bopeto. This means that Bopeto can be used effectively as a preprocessing anomaly detection approach (i.e., Bopeto data can be used to train various anomaly detection models) when there is no indication on contamination ratio in the data set. 

%Although performance degradation is observed not to be a monotonic decreasing function of contamination ratios, in all cases considered here, a model trained on uncontaminated data is assured to outperform its counterpart trained on contaminated versions of the same data. This implies that contamination negatively impacts the performance of the model. Thus, effective anomaly detection should be based on data cleaning or other contamination-robust techniques for more robust models.

We have also conducted experiments on image data sets (MVTec, Visa, MNIST and CIFAR-10) which results are outstanding\textemdash Table \ref{tab:f1}\textemdash and consistent with results on other data sets. The improvement on these image datasets surpasses that observed with the security data sets, because security data sets, are quite challenging. 

Figure \ref{fig:boxplot_bop_difference_all} showcases the enhancements provided by Bopeto across various performance metrics, namely accuracy (Figure \ref{fig:accuracy_boxplot_bop_difference}), precision (Figure \ref{fig:precision_boxplot_bop_difference}), and recall (Figure \ref{fig:recall_boxplot_bop_difference}). Consistent with the findings on F$_1$ score, Bopeto tends to generate data sets with lower contamination levels in most scenarios, thereby enabling unsupervised anomaly detection models to better capture the trends of in-distribution behaviors. However, there are instances where Bopeto's impact on anomaly detection performance is limited, particularly in cases of high contamination. Bopeto may struggle to effectively distinguish between inliers and anomalies during preprocessing when  the contamination ratio exceeds about 23\% of the data set.

\subsection{Bopeto vs IAD and   Bopeto for IAD (Bopeto-IAD) vs IAD}
\begin{figure*}[h]
    \centering
        \includegraphics[width=0.8\textwidth]{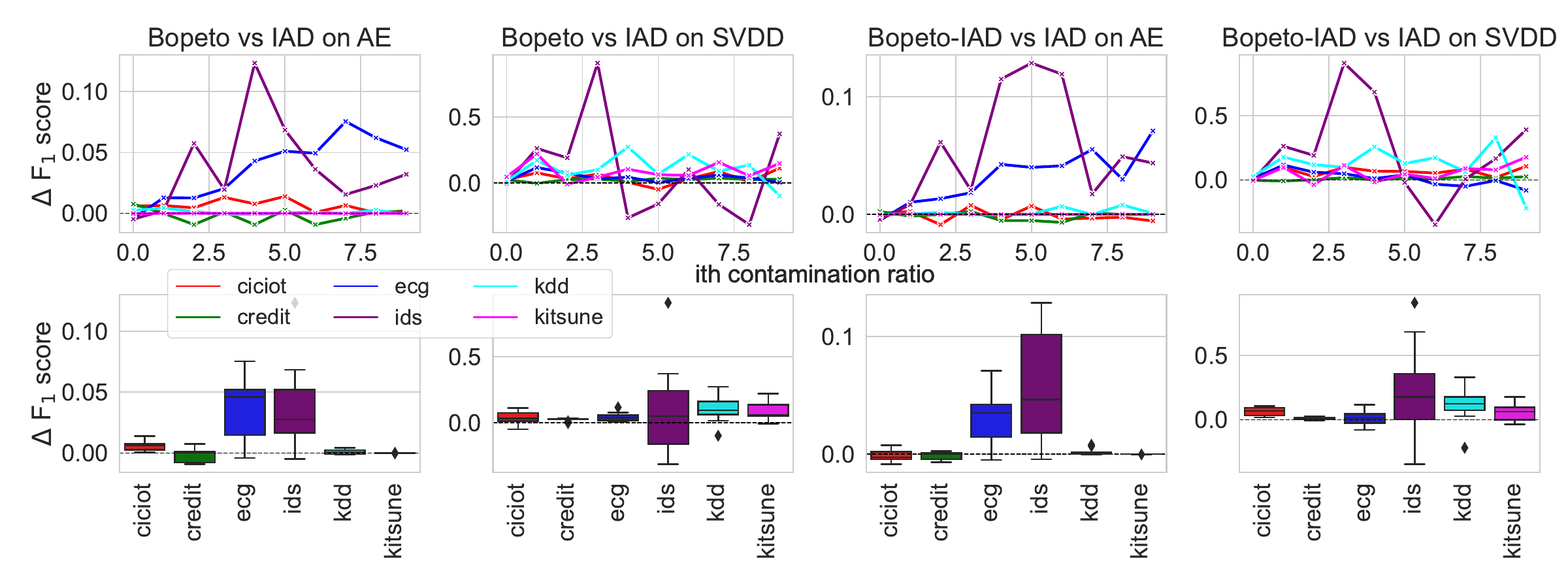}
    \caption{Differences in F$_1$ scores between AE or SVDD trained on Bopeto data using IAD and their counterparts trained on the original data. Positive values indicate the superiority of Bopeto. First row represents detailed $\Delta$ F$_1$ scores per ith contamination ratio (as defined in Figure \ref{fig:dynamics_contamination}) and second row represents distributions of $\Delta$ F$_1$ scores across contamination ratios.}
    \label{fig:f1_iad_difference}
\end{figure*}
Figure  \ref{fig:f1_iad_difference} (first two columns) shows that the impact of Bopeto on AE and SVDD surpasses that of IAD in most cases. IAD assumes that anomalies would exhibit distinct loss values when analyzed individually. However, the main argument of Bopeto is  that distinguishing inliers from anomalies based solely on isolated loss is challenging.  
Instead, by observing their dynamics over iterations, their differentiation becomes clearer. So applying Bopeto prior to IAD can improve model performance.
Figure \ref{fig:f1_iad_difference} (last two columns) indicates that applying IAD on Bopeto data (Bopeto-IAD) has a greater impact on models than using it on the original data. This suggests that Bopeto serves as an effective preprocessing step for handling contaminated data sets using IAD. 
\subsection{Contamination ratio reduction and false positive rates}
\begin{figure*}[h]
\centering
\includegraphics[width=0.8\textwidth]{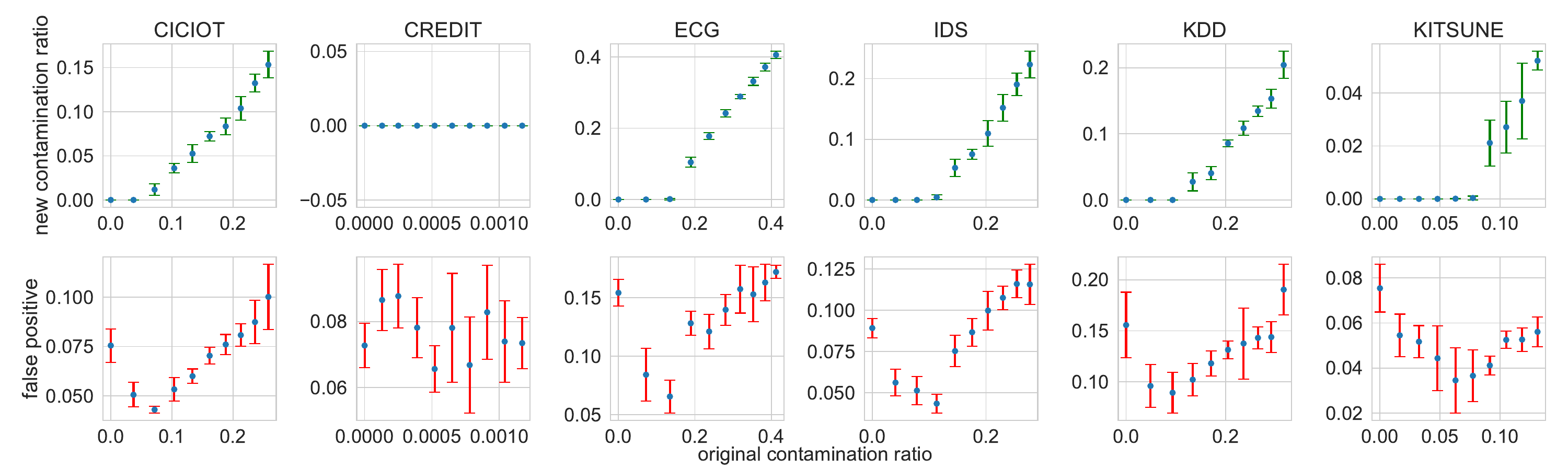}
\caption{5 runs of Bopeto. Contamination reduction (first row) and false positive (second row).}
\label{fig:error_bars}
\end{figure*}
In Figure \ref{fig:error_bars} (first row), narrow error bars are observed in most cases of contamination reduction, indicating low variability. This suggests that Bopeto is quite consistent in contamination reduction. Also, we mentioned earlier that Bopeto might identify some inliers as anomalies. In Figure \ref{fig:error_bars} (second row), the variability in false positives does not align consistently with the original contamination ratio (distribution of false positives can be found in Figure \ref{fig:fp_distribution}). This discrepancy could be attributed to the random threshold fixation applied in IF. Furthermore, it illustrates that Bopeto can identify false anomalies even in uncontaminated data sets. Despite this, false positive rates remain relatively low.

\subsubsection{False positive and false negative in Bopeto}
\begin{figure}[H]
\centering
\includegraphics[width=0.5\textwidth]{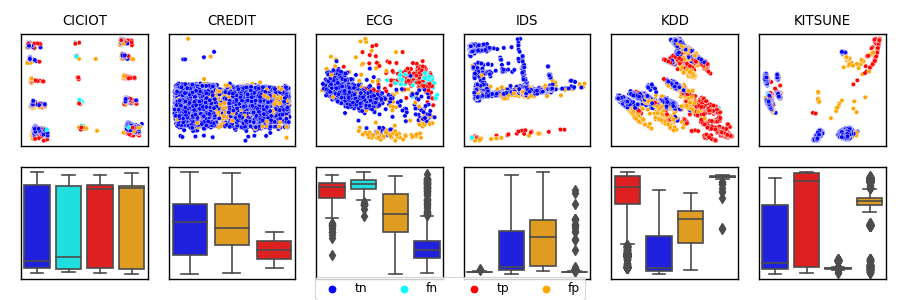}
\caption{Illustrations of where Bopeto fails: tn, tp, fn and fp denote true negative, true positive, false negative and false positive respectively. Anomaly is the positive class here.}
\label{fig:fp_distribution}
\end{figure}
We observed that Bopeto wrongly classifies some inliers as anomalies and fails to recognize some anomalies during cleaning. Figure \ref{fig:fp_distribution} indicates the distributions of data where Bopeto succeeds and, importantly, where it fails. It is clear that most false positives are found either in, or at the edge of, less dense regions. This explains why Bopeto fails to classify them, as they have features similar to anomalies. The impact of their removal depends on their informativeness. An improvement of Bopeto could consist of reducing false positive rates of  informative inliers. %A robust approach to fixing thresholds that can be learned from data distribution would be invaluable.
\subsection{Autoencoder‑agnostic performance analysis}
To assess whether Bopeto is independent of the underlying AE architecture, we conduct a statistical comparison across 3 distinct AE families: a standard AE (which we used in our experiments above), DUAD, and DAGMM. Each architecture is used to generate dynamic‑space representations, on top of which we apply 3 traditional OoD algorithms\textemdash IF, LOF, and OCSVM. This setup allows us to disentangle the contribution of the AE architecture from the behavior of the downstream OoD models.
For every combination of AE architecture and OoD method, we compute both the p‑value and the effect size to quantify the significance and magnitude of performance differences. Table \ref{tab:pvalue_effectsize} summarizes these results, with each cell reporting the p‑value and effect size for a AE–AD pairing.
\begin{table}[ht]
\centering
\scalebox{0.8}{\begin{tabular}{lccc}
\toprule
\textbf{AE Architecture} 
& \textbf{IF} 
& \textbf{LOF} 
& \textbf{OCSVM} \\
\midrule
AE    
& $p = 0.27,\; d = 0.08$ 
& $p = 0.41,\; d = 0.05$ 
& $p = 0.35,\; d = 0.07$ \\
DUAD  
& $p = 0.18,\; d = 0.12$ 
& $p = 0.64,\; d = 0.03$ 
& $p = 0.22,\; d = 0.10$ \\
DAGMM 
& $p = 0.53,\; d = 0.04$ 
& $p = 0.89,\; d = 0.01$ 
& $p = 0.31,\; d = 0.09$ \\
\bottomrule
\end{tabular}}
\caption{Statistical comparison (p-value at 1\% threshold and effect size) between AE architectures and traditional AD methods applied on the dynamic space on F$_1$ scores.}
\label{tab:pvalue_effectsize}
\end{table}

Table \ref{tab:pvalue_effectsize} show that all statistical tests are non‑significant at the 1\% threshold, with consistently small effect sizes across all combinations of autoencoder architectures and traditional anomaly‑detection methods. This pattern indicates that performance differences between AE, DUAD, and DAGMM are negligible when Bopeto is applied to the dynamic space.  %This architecture‑independent behavior is an important property: it confirms that Bopeto captures intrinsic characteristics of the data rather than artifacts introduced by a specific AE model, and it ensures that practitioners can adopt the method without being constrained to a specific AE architecture.

For Bopeto, AE does not need to be fully trained or optimized. We intentionally restrict our analysis to the first ten training iterations, where the model is still unstable and its reconstructions vary significantly from one iteration to the next. This early‑stage variability is precisely what Bopeto exploits: the dynamic sequences capture how each data point responds to these rapid changes in the reconstruction function. What matters, therefore, is not the final accuracy of the autoencoder but the trajectory of its learning process during these initial iterations, where meaningful distinctions between normal and anomalous instances naturally emerge.
\section{Conclusion}
In this paper, we introduced a novel, self-contained data-centric preprocessing framework designed to mitigate the risks associated with training data contamination in unsupervised anomaly detection systems. By analyzing the dynamic trajectory of cosine similarity during the early stages of auto-encoder training, our approach isolates and prunes anomalous training faults without requiring a priori knowledge of the contamination ratio. The resulting ``hard-cleaned'' datasets exhibit a substantial increase in data integrity, leading to up to 100\% contamination reduction. This improvement at the data layer translates directly into enhanced classification performance across diverse downstream detection models, showing an average improvement of 11.2\% in F1-score and 7.3\% in accuracy. Ultimately, our model-agnostic approach provides a practical, low-resource blueprint for engineering dependable security operations\textemdash such as robust network intrusion detection and telemetry monitoring\textemdash in untrusted, contaminated deployment environments.

While our current implementation has proven highly effective for high-dimensional tabular datasets and serialized image representations, it is not currently designed to handle structured graph databases. Given that modern security domains, such as fraud detection and network topology analysis, increasingly rely on graph representations, adapting our dynamic representation space to capture relational and structural dependencies represents a critical avenue for future research. Future work will focus on integrating graph neural networks (GNNs) and neighborhood aggregation dynamics \cite{duan2023graph} into our early-stage training analysis, extending this zero-knowledge data-cleaning paradigm to secure graph-based anomaly detection.

\bibliographystyle{plain}
\bibliography{bibliography}% common bib file

\end{document}